\documentclass[10pt,twocolumn,letterpaper]{article}
\usepackage[pagenumbers]{cvpr} % To force page numbers, e.g. for an arXiv version
\usepackage{amssymb} % for arrow symbols
\usepackage{float} % Add this in the preamble
\definecolor{cvprblue}{rgb}{0.21,0.49,0.74}
\usepackage[pagebackref,breaklinks,colorlinks,allcolors=cvprblue]{hyperref}

\def\paperID{7zraq6lnAH} % *** Enter the Paper ID here
\def\confName{CVPR}
\def\confYear{2025}

\title{Understanding the Capabilities of Molecular Graph Neural Networks in Materials Science Through Multimodal Learning and Physical Context Encoding
}
\author{%
Can Polat$^{1}$, Hasan Kurban$^{2}$\thanks{Corresponding authors. Model and dataset is available at \url{https://github.com/KurbanIntelligenceLab/UnderstandingMultiModalGNNs}.}, Erchin Serpedin$^{1}$, and Mustafa Kurban$^{3*}$\\[1mm]
\footnotesize
$^{1}$Dept. of Electrical and Computer Engineering, Texas A\&M University, College Station, TX 77843, USA\\
\footnotesize
$^{2}$College of Science and Engineering, Hamad Bin Khalifa University, Doha, Qatar\\
\footnotesize
$^{3}$Dept. of Prosthetics and Orthotics, Ankara University, Ankara, Turkey\\[1mm]
\footnotesize
\texttt{\{can.polat, eserpedin\}@tamu.edu, hkurban@hbku.edu.qa, kurbanm@ankara.edu.tr}
}
\usepackage{fancyhdr}

\fancypagestyle{accepted}{
  \fancyhf{}                             % clear all header and footer fields
  \fancyhead[C]{\small Accepted Spotlight Paper at CVPR 2025 for MM4Mat}
}

\begin{document}
\maketitle
\begin{abstract}
Molecular graph neural networks (GNNs) often focus exclusively on XYZ-based geometric representations and thus overlook valuable chemical context available in public databases like PubChem. This work introduces a multimodal framework that integrates textual descriptors—such as IUPAC names, formulas, physicochemical properties, and synonyms--alongside molecular graphs. A gated fusion mechanism balances geometric and textual features, allowing models to exploit complementary information. Experiments on benchmark datasets indicate that adding textual data yields notable improvements for certain electronic properties, though gains remain limited elsewhere. Furthermore, the GNN architectures display similar performance patterns—improving and deteriorating on analogous targets—suggesting they learn comparable representations rather than distinctly different physical insights.
\end{abstract} 
\section{Introduction}
\label{sec:intro}

Computational materials science has traditionally relied on first-principles frameworks such as density functional theory (DFT) \cite{hohenberg1964inhomogeneous} and atomistic simulations—including ab initio molecular dynamics—for elucidating the structures, electronic configurations, and thermodynamic properties of materials. These physically grounded methods have provided deep insights into bonding mechanisms, phase stability, and defect behavior, but their steep computational scaling (often as $O(N^3)$ with system size) and high resource requirements limit studies of large and chemically complex systems \cite{ratcliff2017challenges}. Moreover, with their unimodal outputs—typically restricted to geometric configurations or electronic energy landscapes—conventional techniques overlook complementary chemical metadata and spectroscopic observables that could enrich predictive models \cite{becke2014perspective}.

Density functional tight-binding (DFTB) methods mitigate some of DFT’s computational bottlenecks by employing minimal basis sets and precomputed integrals, enabling simulations of substantially larger systems at a fraction of the cost \cite{liu2019efficient}. However, this efficiency gain comes at the expense of accuracy and transferability: DFTB parameterizations are often tailored to narrow chemical domains and may misrepresent long-range dispersion and many-body polarization effects \cite{goyal2014molecular}. Even full DFT calculations require empirical dispersion corrections to capture van der Waals interactions accurately, and typically struggle with excited-state phenomena or strongly correlated materials without further approximations  \cite{grimme2010consistent}. These trade-offs between cost, accuracy, and scope underscore the need to integrate first-principles outputs into multimodal, data-driven frameworks that can leverage structural, chemical, and spectroscopic information in tandem.

Recent advancements in machine learning (ML) \cite{alpaydin2021machine} have significantly influenced materials science and computational chemistry by enabling surrogate models that learn directly from data rather than relying solely on physics-based simulations \cite{karniadakis2021physics}. Graph neural networks (GNNs) have emerged as a powerful tool for modeling molecular structures due to their ability to capture intricate atomic interactions \cite{wu2020comprehensive}; however, most state-of-the-art approaches depend predominantly on structural information derived from XYZ files, which encode only the geometric and connectivity aspects of molecules \cite{pfau2020ferminet,hu2021forcenet,vonglehn2022psiformer,gupta2024equi}.

To complement geometry‐focused representations, public chemical resources such as PubChem \cite{kim2025pubchem} furnish extensive metadata—including IUPAC names, molecular formulas, computed physicochemical descriptors, and spectral signatures. Incorporating these diverse data modalities into a unified multimodal dataset can provide richer, more contextually grounded molecular representations, thereby boosting the accuracy and robustness of predictive models \cite{baltruvsaitis2018multimodal,liang2024foundations}.

\begin{figure}[h]
    \centering
    \includegraphics[width=1\linewidth]{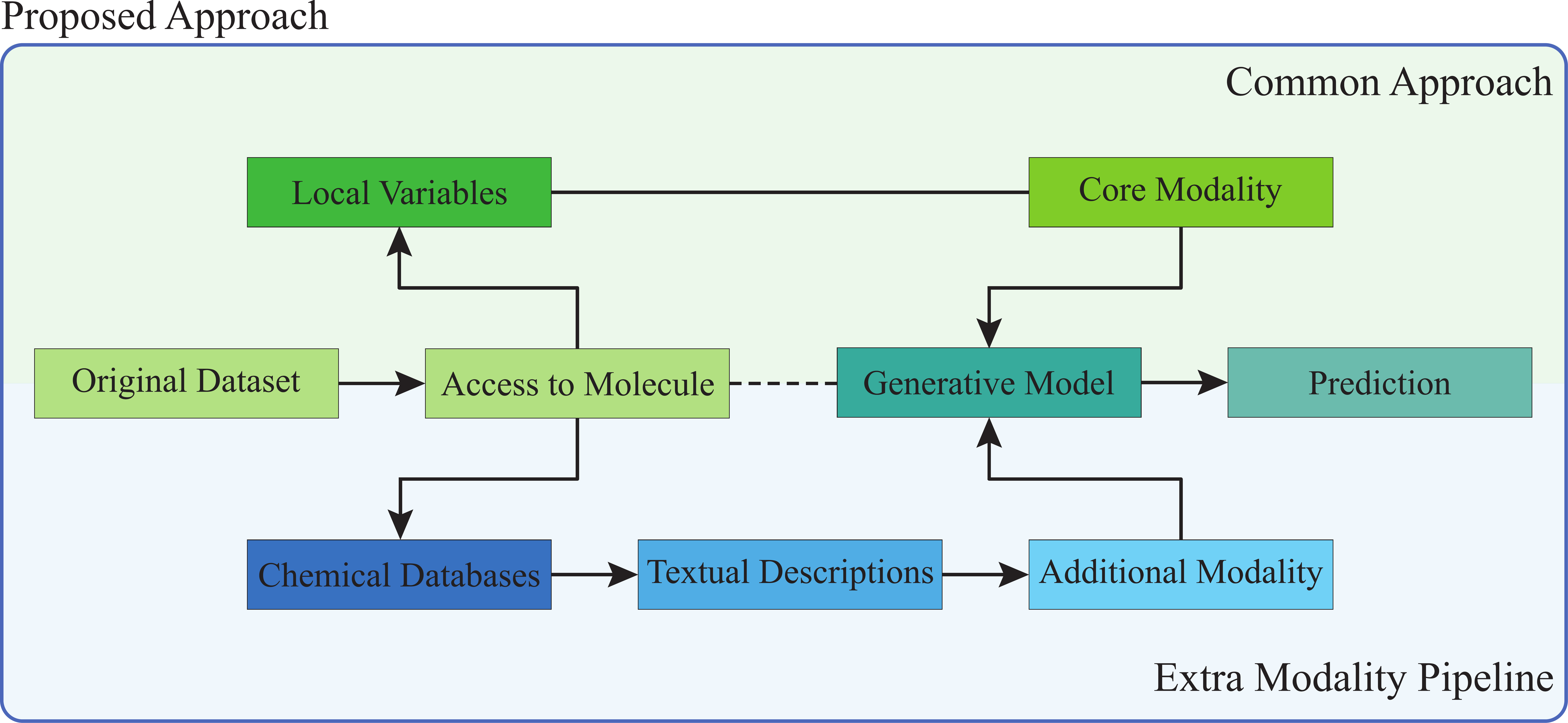}
    \caption{Comparison of common SOTA approaches and the proposed approach. The proposed approach is able to utilize multiple  chemical databases for different property annotations of materials as an additional modality, whereas the common approaches rely solely on XYZ files, which  consist of element names and their 3D coordinates.}
    \label{fig:overallScheme}
\end{figure}

This work introduces a systematic framework for enriching molecular representations with textual annotations from PubChem to enhance the predictive performance of machine learning models across molecular property benchmarks. A gated fusion mechanism \cite{arevalo2020gated} is employed to adaptively combine textual and geometric features, enabling the models to leverage complementary information from both modalities in a balanced and data-driven manner. The key contributions are:  (1) curating and incorporating publicly available textual chemical descriptors into benchmark datasets, creating a novel multimodal resource; (2) extending and evaluating four SOTA geometric deep learning models with text-based features; (3) demonstrating through extensive experiments that incorporating textual descriptors can significantly enhance performance on multiple prediction tasks; and (4) the results indicate that different model architectures consistently leverage textual cues and converge on similar patterns in the data, demonstrating that while textual information enhances predictions, it does not produce fundamentally distinct learned representations across architectures. The general scheme of the process is shared in Figure \ref{fig:overallScheme}.

\section{Related Work}
\subsection{Quantum Mechanical Foundations and Density Functional Approximations}
The many-electron problem is fundamentally described by the Schrödinger equation (SE):
\begin{equation}
\hat{H} \psi(\mathbf{r}_1, \ldots, \mathbf{r}_N) = E\,\psi(\mathbf{r}_1, \ldots, \mathbf{r}_N),
\label{eq:se}
\end{equation}
with the Hamiltonian operator given by
\begin{equation}
\hat{H} = -\sum_{i=1}^{N} \frac{1}{2}\nabla_i^2 - \sum_{i=1}^{N}\sum_{A=1}^{M}\frac{Z_A}{|\mathbf{r}_i - \mathbf{R}_A|} + \sum_{i=1}^{N}\sum_{j>i}\frac{1}{|\mathbf{r}_i-\mathbf{r}_j|}.
\label{eq:hamiltonian}
\end{equation}
Due to the exponential increase in complexity with the number of particles \cite{sherrill1999configuration, helgaker2013molecular}, approximate methods become essential. DFT offers a practical solution by recasting the problem in terms of electron density. Within the Kohn-Sham framework, the total energy is modeled as:
\begin{equation}
E_\text{total}[\rho(\mathbf{r})] = T_s[\rho] + \int v_\text{ext}(\mathbf{r})\,\rho(\mathbf{r})\,d\mathbf{r} + E_H[\rho] + E_\text{xc}[\rho],
\label{eq:KS}
\end{equation}
where \(T_s[\rho]\) represents the kinetic energy of a fictitious system of non-interacting electrons, \(v_\text{ext}\) denotes the external potential, \(E_H[\rho]\) is the classical electrostatic energy, and \(E_\text{xc}[\rho]\) encapsulates the exchange-correlation effects. Although DFT achieves high accuracy, its computational complexity—approximately \(O(n^3T)\) with respect to the number of electrons (or basis functions) \(n\) and self-consistent iterations \(T\) \cite{dawson2022density}—renders it less feasible for large datasets.

\subsection{Machine Learning in Materials Science}

Neural networks have emerged as powerful tools for approximating solutions to the many-body SE, mapping atomic configurations to electronic structures with high fidelity while enabling efficient, scalable predictions of energy levels, physical observables, and wave functions, significantly reducing computational costs in quantum simulations \cite{manzhos2009improved, cai2018approximating, hermann2020deep, mianroodi2021teaching, schleder2019dft, jinnouchi2017predicting, snyder2012finding, snyder2013orbital, li2016pure, kurban2024enhancing}. Recent advancements, such as physics-informed neural networks, further enhance model generalization by embedding governing equations into the training process, improving data efficiency, reducing reliance on large labeled datasets, and bridging traditional ab initio methods with deep learning to yield more interpretable and reliable electronic structure predictions \cite{raissi2019physics, samaniego2020energy, jiang2022physics, chen2020physics, rasp2021data, kashinath2021physics}.

ML models in molecular property prediction have leveraged a diverse array of architectures—ranging from GNNs and transformer‐based models to explicitly equivariant frameworks—to encode domain‐specific inductive biases and improve generalization across chemical systems. Methods such as DTNN \cite{schutt2017quantumdtnmn}, FermiNet \cite{pfau2020ab}, SpookyNet \cite{unke2021spookynet}, and QuantumShellNet \cite{polat2025quantumshellnet} demonstrated the power of rotational equivariance, while more recent approaches like Equiformer-v2 \cite{liao2023equiformerv2}, Pure2DopeNet \cite{polat2024multimodal}, and BLMM \cite{wei2024crystal} have shown that self-attention mechanisms and latent-space regularization can further refine electronic property predictions. Complementing these developments, models such as TorchMD-Net \cite{torchMD} and CGNN \cite{cgnn}, which treat atoms and their local environments as nodes and edges, now set the benchmark on molecular dynamics datasets like MD17 \cite{md17} and on bulk crystalline property tasks derived from the Materials Project database \cite{jain2020materials}. Equivariant architectures—including E(n)-Equivariant GNNs \cite{satorras2021n}, SE(3)-Transformers \cite{fuchs2020se}, PaiNN \cite{painn}, SphereNet \cite{spherenet}, and GotenNet \cite{gotennet}—explicitly enforce three-dimensional symmetries, delivering enhanced accuracy and data efficiency when predicting energies, forces, and lattice parameters directly from atomic coordinates.

Standardized datasets have also played a pivotal role in ML driven materials science. Datasets such as QM7 \cite{blum,rupp}, OC20 \cite{chanussot2021open}, and OC22 \cite{tran2023open} serve as essential benchmarks for evaluating generalization and robustness. Meanwhile, multimodal datasets—including ScienceQA \cite{lu2022learn}, SciBench \cite{wang2023scibench}, TDCM25 \cite{polat2025tdcm}, and LabBench \cite{laurent2024lab}—have extended evaluation paradigms to complex, integrative tasks, establishing a rigorous framework for assessing AI-driven scientific discovery.
\begin{figure*}[htbp]
    \centering
    \includegraphics[width=1\linewidth]{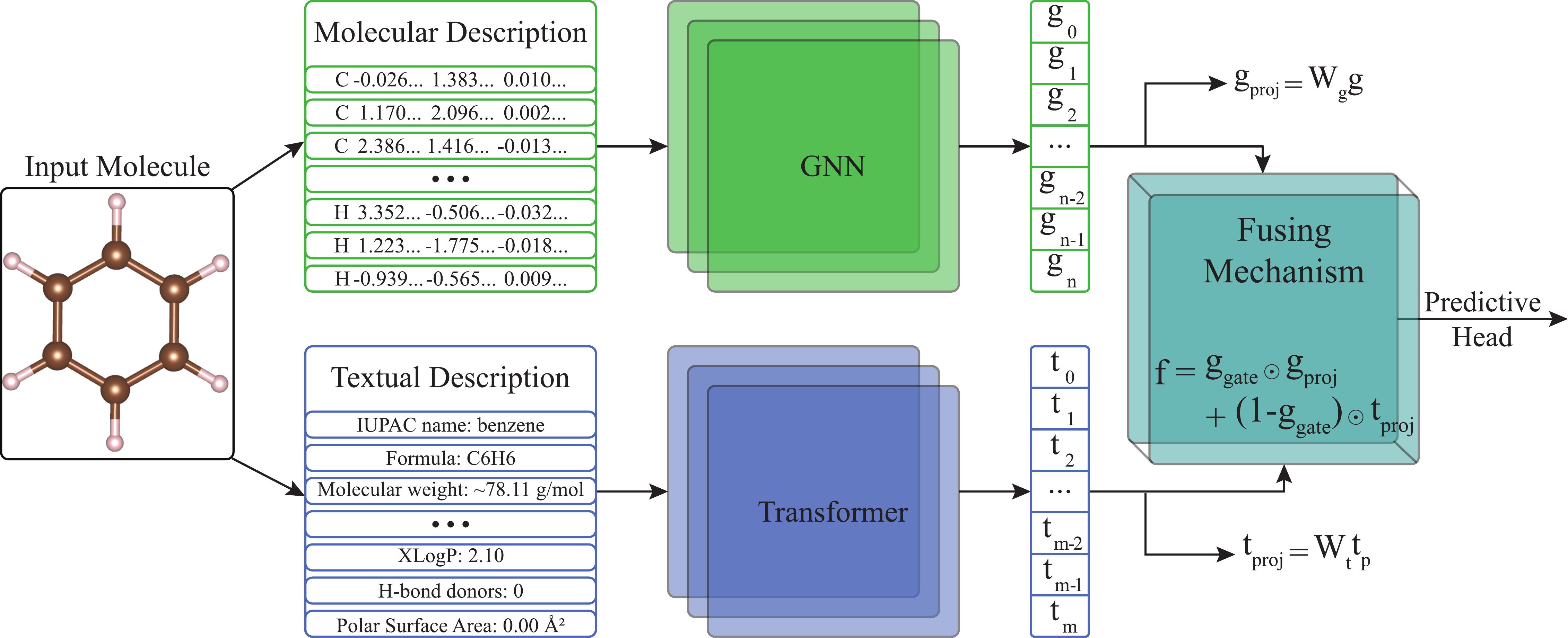}
    \caption{The proposed approach integrates textual descriptions retrieved from the PubChem library with the XYZ structural data provided by the QM9 dataset. Textual descriptions are processed by a transformer model to generate textual embeddings, while XYZ coordinates are input into a GNN to derive molecular embeddings. Subsequently, these embeddings are fused into a unified representation, which is then used to predict molecular properties.}
    \label{fig:our_method}
\end{figure*}

\section{Method}
%Our multimodal pipeline consists of two primary components. First, textual descriptions extracted from the enriched QM9 dataset \cite{ramakrishnan2014quantum,ruddigkeit2012enumeration} are processed using HuggingFace's \cite{wolf2019huggingface} CLIP model \cite{clip} and tokenizer to generate 768-dimensional text embeddings. These embeddings, denoted by $\mathbf{t} \in \mathbb{R}^{768}$, capture semantic and chemical information that goes beyond the basic three-dimensional geometry.

The proposed multimodal pipeline integrates textual and geometric molecular representations through two key components. First, text embeddings are extracted from enriched QM9 dataset descriptions \cite{ramakrishnan2014quantum,ruddigkeit2012enumeration} using HuggingFace’s CLIP model \cite{clip}, yielding a 768-dimensional vector $\mathbf{t}\in\mathbb{R}^{768}$ that captures semantic and chemical context beyond raw geometry. Second, molecular geometries are processed via a message-passing network that learns atomistic interactions and equivariance through iterative neighborhood aggregation. Structural features are encoded via GNNs operating on the molecular point cloud and its connectivity: atoms become nodes initialized with one-hot elemental embeddings; edges connect all atom pairs within a fixed cutoff and are labeled by interatomic distances $d_{ij}$, which are expanded into learnable radial basis functions (e.g., Gaussian or Bessel functions) to enable continuous-filter convolutions modulated by $d_{ij}$ during message updates. Angular relationships between triplets of atoms are captured by expanding both distances and angles into spherical harmonics and radial basis functions, producing directional messages that encode three-body geometry. Higher-order tensor features and learnable tensor products—built from irreducible representations of the rotation group—further enrich each message with fully equivariant geometric context. After $T$ message-passing iterations, the resulting node features $\mathbf{h}_i^{(T)}$ are globally pooled (by sum or gated pooling) into a fixed-size geometry embedding $\mathbf{g}\in\mathbb{R}^n$. 

All geometric encodings satisfy the fundamental molecular symmetries by design: only relative coordinates $\mathbf{r}_{ij}=\mathbf{r}_j-\mathbf{r}_i$ enter the filter functions, ensuring translation invariance; neighborhood aggregation is a symmetric operation, guaranteeing permutation invariance under any reordering of atom indices; and rotational symmetry is upheld by using invariant distance inputs for scalar features, applying equivariant spherical-basis expansions that rotate covariantly, and contracting or pooling these into invariants as required, thereby enforcing both rotation equivariance and invariance throughout the network.

To integrate textual and structural molecular representations, a gated fusion mechanism is employed to dynamically balance contributions from both modalities. The text embedding $\mathbf{t}$ is first projected into a lower‐dimensional space via a learnable projection head:
\begin{equation}
P: \mathbb{R}^{768} \to \mathbb{R}^{d},
\end{equation}
yielding a compact text representation:
\begin{equation}
\mathbf{t}_p = P(\mathbf{t}),
\end{equation}
where $d$ is the reduced embedding dimension (e.g., $d = 16$). Independently, the graph‐level structural embedding $\mathbf{g}\in\mathbb{R}^n$ produced by the GNN is linearly transformed via
\begin{equation}
\mathbf{g}_{\mathrm{proj}} = W_g\,\mathbf{g}, 
\quad
\mathbf{t}_{\mathrm{proj}} = W_t\,\mathbf{t}_p,
\end{equation}
where $W_g\in\mathbb{R}^{n\times n}$ and $W_t\in\mathbb{R}^{n\times d}$ are learnable matrices that align both modalities into the same latent space. To ensure that structural and textual embeddings are directly comparable in magnitude, each projected vector is passed through layer normalization:
\begin{equation}
\tilde{\mathbf{g}} = \mathrm{LayerNorm}(\mathbf{g}_{\mathrm{proj}}), 
\quad
\tilde{\mathbf{t}} = \mathrm{LayerNorm}(\mathbf{t}_{\mathrm{proj}}).
\end{equation}
A gating vector is then computed over the normalized features:
\begin{equation}
\mathbf{g}_{\mathrm{gate}} = \sigma\Bigl(W\,\bigl[\tilde{\mathbf{g}}\;\|\;\tilde{\mathbf{t}}\bigr] + b\Bigr)\in\mathbb{R}^n,
\end{equation}
where $W\in\mathbb{R}^{(n+d)\times n}$ and $b\in\mathbb{R}^n$ are learnable, and $\sigma(\cdot)$ is the sigmoid activation. The final fused representation $\mathbf{f}\in\mathbb{R}^n$ is obtained by modality‐selective interpolation:
\begin{equation}
\mathbf{f} = \mathbf{g}_{\mathrm{gate}}\odot\tilde{\mathbf{g}} \;+\;(1 - \mathbf{g}_{\mathrm{gate}})\odot\tilde{\mathbf{t}},
\end{equation}
where $\odot$ denotes the Hadamard (element‐wise) product. All projection and gating parameters are trained end‐to‐end on the target property loss, allowing the network to automatically calibrate any residual scale differences without manual weighting. Finally, $\mathbf{f}$ is passed through fully connected layers to predict the desired material properties. An overview of the proposed scheme is shown in Figure \ref{fig:our_method}.

%While incorporating CLIP-based textual descriptors enhances performance, the observed similarities across different multimodal architectures suggest a capacity bottleneck inherent in the underlying message-passing design. This highlights a key challenge: improving structural representations remains crucial for further advancing multimodal materials discovery.

\begin{table*}[htbp]
    \centering
    \begin{tabular}{lcccc}
        \toprule
        Property (Target)& SchNet & DimeNet++ & Equiformer & FAENet \\
        \midrule
         Dipole Moment            & \textcolor{green}{12.07\% $\uparrow$} & \textcolor{green}{8.05\% $\uparrow$}  & \textcolor{green}{7.17\% $\uparrow$}  & \textcolor{red}{1.38\% $\downarrow$} \\
         Isotropic Polarizability  & \textcolor{red}{14.60\% $\downarrow$}  & \textcolor{red}{1.82\% $\downarrow$}   & \textcolor{red}{27.98\% $\downarrow$} & \textcolor{green}{8.06\% $\uparrow$}  \\
         HOMO                        & \textcolor{green}{20.36\% $\uparrow$}  & \textcolor{green}{23.63\% $\uparrow$}  & \textcolor{green}{19.65\% $\uparrow$}  & \textcolor{green}{9.00\% $\uparrow$}  \\
         LUMO                       & \textcolor{green}{15.42\% $\uparrow$}  & \textcolor{green}{12.22\% $\uparrow$}  & \textcolor{green}{4.92\% $\uparrow$}   & \textcolor{green}{1.07\% $\uparrow$}  \\
         HOMO-LUMO Gap                        & \textcolor{green}{14.82\% $\uparrow$}  & \textcolor{green}{19.47\% $\uparrow$}  & \textcolor{green}{12.95\% $\uparrow$}  & \textcolor{green}{6.50\% $\uparrow$}  \\
         Electronic Spatial Extent & \textcolor{red}{3.34\% $\downarrow$}   & \textcolor{green}{4.04\% $\uparrow$}   & \textcolor{red}{3.97\% $\downarrow$}  & \textcolor{green}{12.35\% $\uparrow$} \\
         ZPVE                       & \textcolor{green}{0.88\% $\uparrow$}   & \textcolor{red}{12.12\% $\downarrow$}  & \textcolor{red}{5.71\% $\downarrow$}  & \textcolor{red}{28.21\% $\downarrow$} \\
        \bottomrule
    \end{tabular}
        \caption{Percentage change in MAE for multimodal models incorporating textual chemical descriptors compared to their geometric-only counterparts on the QM9 dataset. Results are averaged over three independent runs. Green upward arrows indicate improved performance (lower MAE), while red downward arrows denote deterioration (higher MAE).}
    \label{tab:multimodal_changes_full}
\end{table*}

\section{Experiments}
\subsection{Dataset}

The QM9 dataset is a widely used benchmark in quantum chemistry, comprising approximately 134,000 small organic molecules represented by their XYZ coordinates. While these files encode molecular geometry, they lack key chemical descriptors that could enhance predictive modeling. To address this limitation, QM9 is augmented with textual descriptors from PubChem, including the IUPAC name, molecular formula, molecular weight, XLogP, hydrogen bond donor and acceptor counts, rotatable bond count, topological polar surface area, formal charge, synonyms, and spectral data. This multimodal enrichment provides complementary chemical context, extending beyond structural information to improve downstream property prediction and materials discovery. Only molecules with available PubChem descriptors are included from the final annotated dataset.

% Table~\ref{tab:example_descriptors} shows two example entries with enhanced chemical descriptors obtained through the annotation process. These descriptors complement the XYZ geometry in QM9 by adding critical chemical information that can improve property predictions.

% \begin{table*}[ht]
% \centering
% \begin{tabular}{lcc}
% \hline
% \textbf{Property} & \textbf{Example 1} & \textbf{Example 2} \\
% \hline
% IUPAC Name & methane & oxidane \\
% Molecular Formula & CH\(_4\) & H\(_2\)O \\
% Molecular Weight & $\sim$16.04 g/mol & $\sim$18.02 g/mol \\
% XLogP & 0.60 & -0.50 \\
% H-bond Donors & 0 & 1 \\
% H-bond Acceptors & 0 & 1 \\
% Rotatable Bonds & 0 & 0 \\
% Topological Polar Surface Area & 0.00 Å\(^2\) & 1.00 Å\(^2\) \\
% Synonyms (partial list) & methane; Methyl hydride; Marsh gas & water; 7732-18-5; Distilled water \\
% \hline
% \end{tabular}
% \caption{Annotated chemical descriptors for two representative molecules. These enhanced annotations provide detailed chemical context that supplements the geometric data in QM9, thereby aiding in more accurate property prediction.}
% \label{tab:example_descriptors}
% \end{table*}

\subsection{Results}

The implementations of the dataset and the models are done by using PyTorch \cite{pytorch} and PyTorch-Geometric \cite{torch_geo} libraries in addition to the official repositories. Experimental evaluation employed three-fold cross-validation with a batch size of 64, a learning rate of $1\times10^{-3}$; model parameters were optimized using the Adam optimizer and the mean absolute error loss was used as training criterion. The work evaluate the impact of integrating textual chemical descriptors with geometric data using a diverse set of SOTA architectures in molecular property prediction: SchNet \cite{schutt2017schnet}, DimeNet++ \cite{gasteiger2020fast}, Equiformer \cite{liao2022equiformer}, and FAENet \cite{duval2023faenet}. SchNet employs a continuous-filter convolutional framework for robust geometric processing, while DimeNet++ enhances directional message passing with spherical harmonics. Equiformer leverages equivariant neural networks to enforce physical symmetries, and FAENet incorporates frame averaging with flexible skip connections to capture rich atom-level features. Together, these models provide a comprehensive benchmark for evaluating the benefits of multimodal molecular representations.

Table~\ref{tab:multimodal_changes_full} summarizes the percentage changes in mean absolute error (MAE) across three independent runs, comparing geometric-only models to their multimodal counterparts. The evaluated targets include dipole moment, isotropic polarizability, highest occupied molecular orbital (HOMO) energy, lowest unoccupied molecular orbital (LUMO) energy, HOMO-LUMO gap, electronic spatial extent, and zero point vibrational energy (ZPVE). Positive values (green, upward arrows) indicate better performance (lower MAE), while negative values (red, downward arrows) indicate deterioration.

Integrating textual descriptors generally enhances predictions for dipole moment, HOMO energy, and HOMO-LUMO gap, with SchNet, DimeNet++, and Equiformer showing the most substantial gains. This suggests that molecular formula, weight, and other chemical context complement geometric data, particularly for electronic properties.

Conversely, isotropic polarizability sees a decline in accuracy across most models except FAENet, indicating that textual descriptors may not effectively capture its underlying physics. The results for electronic spatial extent and ZPVE vary across architectures, highlighting the selective impact of multimodal features—properties closely tied to chemical identity and electronic structure benefit the most, whereas those governed by complex many-body interactions do not consistently improve.

For energetic properties—including internal energy at 0 K and 298.15 K, enthalpy, free energy, and heat capacity at 298.15 K—the performance of all models deteriorates upon introduction of textual descriptors. This suggests that the additional descriptors do not provide complementary information relevant to these predictions, highlighting the necessity for more task-specific or physics-informed descriptors to achieve accurate modeling in this domain.

\section{Discussion and Conclusion}

The results show that integrating textual descriptors with geometric molecular data enhances predictive performance for dipole moment, frontier orbital energies, and the HOMO–LUMO gap. However, gains are inconsistent across properties; isotropic polarizability and ZPVE exhibit limited or negative effects, indicating that textual features do not universally capture underlying chemical complexities.

Notably, multimodal extensions of SOTA models yield similar performance trends, suggesting that existing architectures extract overlapping information and are inherently constrained in leveraging textual cues. Textual descriptors prove most beneficial for properties closely tied to chemical identity and electronic structure but contribute less to those governed by complex many‐body interactions.

These findings highlight the need for more expressive descriptors and advanced fusion mechanisms, such as attention‐based or dynamically gated architectures, to better integrate multimodal information. Refining descriptor selection and exploring alternative embedding techniques could further enhance predictive accuracy, particularly for properties where textual augmentation currently provides minimal or negative improvements.

Furthermore, although the fused model outperforms geometry‐only baselines on many targets, the incorporation of textual embeddings unexpectedly degrades performance on properties such as isotropic polarizability and ZPVE. This negative transfer can be attributed to three main factors: modality irrelevance, gradient conflicts, and distributional mismatch. Generic textual descriptors often lack the fine‐grained quantum‐mechanical detail required for properties driven by electron‐density distributions, effectively acting as noise in the fused representation. Joint training on uncorrelated modalities can induce gradient conflicts in which one modality’s gradients dominate and steer the model away from an optimum for the other, while misalignment between textual pretraining corpora and downstream property distributions can cause the model to overfit irrelevant text patterns. Recognizing these factors underscores the importance of mitigation strategies—such as modality‐specific normalization, gradient surgery to resolve conflicts, and curriculum training to align modality distributions—to preserve unimodal performance and harness the full potential of multimodal fusion.

\small
\bibliographystyle{ieeetr}
\bibliography{main}

\end{document}